\RequirePackage{fix-cm}
\documentclass[sn-mathphys-num, iicol]{sn-jnl}

\usepackage{graphicx}%
\usepackage{multirow}%
\usepackage{amsmath,amssymb,amsfonts}%
\usepackage{amsthm}%
\usepackage{mathrsfs}%
\usepackage[title]{appendix}%
\usepackage{xcolor}%
\usepackage{textcomp}%
\usepackage{manyfoot}%
\usepackage{booktabs}%
\usepackage{algorithm}%
\usepackage{algorithmicx}%
\usepackage{algpseudocode}%
\usepackage{listings}%
\usepackage[capitalize]{cleveref}

\usepackage{bm}
\usepackage{multirow}
\usepackage{makecell}
\usepackage{subcaption}
\usepackage{enumitem} 

\hypersetup{bookmarksdepth=3}

\DeclareMathAlphabet{\pazocal}{OMS}{zplm}{m}{n}

\newcommand{\mat}[0]{\begin{bmatrix}}
\newcommand{\mate}[0]{\end{bmatrix}}

\newcommand{\vq}{\mathbf{q}}

\newcommand{\vx}{\mathbf{x}}

\newcommand{\cF}{\mathcal{F}}

\newcommand{\cK}{\mathcal{K}}

\newcommand{\cM}{\mathcal{M}}

\newcommand{\cV}{\mathcal{V}}
\newcommand{\cW}{\mathcal{W}}

\newcommand{\R}{\mathbb{R}}

\makeatletter

\newcommand*\wthelper[2]{%
        \hbox{\dimen@\accentfontxheight#1%
                \accentfontxheight#11.3\dimen@
                $\m@th#1\widetilde{#2}$%
                \accentfontxheight#1\dimen@
        }%
}

\newcommand*\accentfontxheight[1]{%
        \fontdimen5\ifx#1\displaystyle
                \textfont
        \else\ifx#1\textstyle
                \textfont
        \else\ifx#1\scriptstyle
                \scriptfont
        \else
                \scriptscriptfont
        \fi\fi\fi3
}
\makeatother

\begin{document}

\title[STEM: Semantic Target Search and Exploration using MAVs in Cluttered Environments]{STEM: Semantic Target Search and Exploration using MAVs in Cluttered Environments}

\author[1]{\fnm{Nikhil} \sur{Sethi}}\email{sethi.nirvil@gmail.com}
\equalcont{These authors contributed equally to this work.}

\author*[1]{\fnm{Max} \sur{Lodel}}\email{max.lodel@gmail.com}
\equalcont{These authors contributed equally to this work.}

\author[1]{\fnm{Laura}\sur{Ferranti}}\email{l.ferranti@tudelft.nl}

\author[1,2]{\fnm{Robert}\sur{Babu\v{s}ka}}\email{r.babuska@tudelft.nl}

\author[1]{\fnm{Javier}\sur{Alonso-Mora}}\email{j.alonsomora@tudelft.nl}

\affil*[1]{\orgdiv{Department of Cognitive Robotics}, \orgname{Delft University of Technology},
\orgaddress{\country{Netherlands}}}

\affil[2]{\orgdiv{CIIRC}, \orgname{Czech Technical University in Prague},
\orgaddress{\country{Czech Republic}}}

\abstract{
Autonomous target search is crucial for deploying Micro Aerial Vehicles (MAVs) in emergency response and rescue missions. 
Existing approaches either focus on 2D semantic navigation in structured environments -- which is less effective in complex 3D settings, or on robotic exploration in cluttered spaces -- which often lacks the semantic reasoning needed for efficient target search. 
This paper overcomes these limitations by proposing a novel framework that utilizes 
a semantically-guided viewpoint planner
to minimize target search and exploration time in unstructured 3D environments using an MAV. 
Specifically, we develop a combinatorial planner that generates efficient semantic exploration plans by prioritizing viewpoints that likely lead to the target. 
To guide the planner towards the target, an active perception pipeline is developed that propagates semantic priorities of observed objects into neighboring frontier voxels for computing semantic information gains of frontier viewpoints.
In addition, we demonstrate how LLM-based similarity scores can be leveraged as semantic priority input to our pipeline.
Evaluations in two distinct simulation environments show that the proposed method consistently outperforms baselines by quickly finding the target while maintaining reasonable exploration times. Real-world experiments with an MAV further demonstrate the method's ability to handle practical constraints like limited battery life, small sensor range, and semantic uncertainty.}


\maketitle

\section{Introduction}

Micro Aerial Vehicles (MAVs) are a promising tool to effectively search and explore complex, unknown environments in 
domains such as search and rescue, inspection, and environmental monitoring.
To relieve human operators from the challenging task of guiding MAVs through uncertain environments, methods for autonomous search of target objects are critical to improve the efficiency and effectiveness of such MAV missions.

In large and cluttered environments with many occlusions, the time efficiency of the search strategy is crucial due to limited available flight time and onboard sensor capabilities.
Humans can efficiently search unknown spaces by leveraging their experience as well as semantic information, such as observed objects, to reason about the target's likely location. 
For example, dangerous chemicals are more likely to be found in a storage room than in an office.
Building on this idea,
recent work \cite{chaplot_semantic_2020,kim2022,ramakrishnan2022a,hahn2021,yokoyama_vlfm_2023,chen_how_2023,gadreCoWsPastureBaselines2022,huang2022,ginting_seek_2024}
has shown that learning such semantic priors can significantly reduce the target search time by guiding the robot toward promising regions.
However, as semantic observations or priors can be uncertain or unavailable in many real-world scenarios, purely relying on such semantic guidance may lead to inefficient behavior.
Therefore, a robust target search strategy should balance semantic search and coverage-maximizing exploration to ensure an efficient and successful search.

Existing works on semantically-guided target search \cite{chaplot_semantic_2020,kim2022,ramakrishnan2022a,hahn2021,yokoyama_vlfm_2023,chen_how_2023,gadreCoWsPastureBaselines2022,huang2022,ginting_seek_2024} have focused on ground robots moving in 2D, while MAVs' 3D capabilities remain underexplored.
In particular, MAVs can overcome occlusions in cluttered environments by changing their altitude and, therefore, improve search efficiency.
Moreover, these target search methods rely on inferring semantic relationships from large pre-trained models \cite{yokoyama_vlfm_2023}, 
which might require fine-tuning for highly specific scenarios.
This underlines the need for integrating and balancing semantic search with efficient coverage-maximizing exploration approaches such as \cite{zhou2021fuel,cao2023tare,meng_two-stage_2017}.
These methods achieve high coverage efficiency by leveraging long-horizon combinatorial planning techniques, but ignore semantic information that could guide the search towards the target.

In this work, we present STEM, a framework for \textbf{S}emantic \textbf{T}arget Search and \textbf{E}xploration for \textbf{M}AVs.
By building on recent advances in exploration planning and semantics-driven navigation, our framework enables both semantically-guided and efficient exploration in complex 3D environments.

\subsection{Related Work}\label{sec:related_work}

In this section, we discuss existing approaches and how they relate to our work, starting with planning approaches for pure coverage exploration, and then focusing on target search approaches that leverage semantic information.

\subsubsection{Exploration Planning}
The problem of navigating a robot autonomously through an unknown environment to build a complete map from sensor observations has been investigated using a variety of different approaches.
The fundamental idea of most approaches is to choose robot actions or plans such that future \textit{viewpoints} efficiently 
minimize unknown space in the environment. Viewpoints are poses in 3D space from which the robot can observe the environment.

Sampling-based approaches randomly sample viewpoints in free space and evaluate their potential information gain about the map.
Rapidly Exploring Random Trees (RRT) have been used to plan a local tree of informative viewpoints \cite{bircher_receding_2016,dharmadhikariMotionPrimitivesbasedPath2020,schmid_efficient_2020,dang_visual_2018},
with \cite{dang_visual_2018} integrating object search by prioritizing salient objects in the environment.
However, the high computational costs limit the planning horizon, leading to greedy and inefficient exploration.

Frontier-based exploration methods focus on observing the boundary between known and unknown space, the \textit{frontiers}, to incrementally reduce the unknown space. 
While early methods \cite{yamauchi1997frontier} just choose the closest frontier as the next observation target, recent work shows that selecting frontiers to maintain high flight speed improves exploration efficiency \cite{cieslewski_rapid_2017}. 
Similarly to sampling-based approaches, these methods lack long-horizon planning capabilities.

Recent works have shown that combining elements of both sampling-based and frontier-based exploration with planning can improve exploration efficiency.
The approaches in \cite{zhou2021fuel,cao2023tare,huang_fael_2023,meng_two-stage_2017,zhangFALCONFastAutonomous2024}
sample viewpoints around different frontiers, and then find a time-optimal global plan that connects these viewpoints using combinatorial planning.
Following this approach, \cite{meng_two-stage_2017,cao2023tare,zhou2021fuel} plan over frontier viewpoints by solving a Travelling Salesman Problem (TSP). Specifically, the FUEL framework \cite{zhou2021fuel} considers drone dynamics in the TSP cost matrix to achieve efficient and agile exploration.
In FAEL \cite{huang_fael_2023}, a version of the minimum latency problem is used for planning that prioritizes frontier viewpoints with high coverage gains.
Such approaches can be scaled to large environments using coarse global planning \cite{cao2023tare,zhangFALCONFastAutonomous2024}.

The authors of FUEL \cite{zhou2021fuel} show that their approach achieves efficient and robust exploration performance on a real-world MAV platform in varying, complex 3D environments, due to an effective integration of global and local planning.
As we are interested in 3D semantically-guided exploration with MAVs, we build on the FUEL framework \cite{zhou2021fuel} as an exploration baseline, extending it with 3D semantic representations and planning to enable semantic target search. 
Our target search planner uses an MLP-based formulation similar to FAEL \cite{huang_fael_2023} and integrates semantic information to guide exploration toward target-relevant objects.

\subsubsection{Target Search}

Autonomously searching for a target object in an unknown environment has been primarily investigated in the domain of indoor structured environments such as apartments, where clear semantic relationships between objects exist \cite{chaplot_semantic_2020,kim2022,ramakrishnan2022a,hahn2021,yokoyama_vlfm_2023,chen_how_2023,gadreCoWsPastureBaselines2022,huang2022,ginting_seek_2024}.

These approaches differ in the source of learned semantic priors and the planning strategy used to guide the robot toward the target.
In earlier works, domain-specific environment datasets are used for training navigation policies using reinforcement learning (RL) \cite{chaplot_semantic_2020,kim2022} and training cost-to-go functions using self-supervised learning \cite{ramakrishnan2022a,hahn2021}.
Conversely, recent works \cite{yokoyama_vlfm_2023,chen_how_2023,ginting_seek_2024,gadreCoWsPastureBaselines2022,huang2022} 
use foundation models such as Vision-Language Models (VLM) \cite{radford2021learning,liBLIP2BootstrappingLanguageImage2023} or Large Language Models (LLM) \cite{devlin2018bert}
trained on internet-scale data.
The works \cite{yokoyama_vlfm_2023,chen_how_2023,gadreCoWsPastureBaselines2022,huang2022} demonstrate zero-shot VLM/LLM-based target search in indoor environments using embedding similarity scores to choose exploration frontiers. In \cite{yokoyama_vlfm_2023,gadreCoWsPastureBaselines2022,huang2022}, the frontier selection is facilitated by propagating similarity scores into 2D \cite{yokoyama_vlfm_2023,gadreCoWsPastureBaselines2022} or 3D \cite{huang2022} map representations.
SEEK \cite{ginting_seek_2024} proposes to distill semantic knowledge from an LLM into a lightweight model for efficient online inference. 

The planning strategies used in most of these works are either based on learned navigation policies \cite{chaplot_semantic_2020,kim2022} or greedy frontier selection \cite{ramakrishnan2022a,hahn2021,yokoyama_vlfm_2023,chen_how_2023}. In contrast, \cite{ginting_seek_2024} uses a Bayesian network prediction model and value iteration planning to choose the best region to search.

The main limitation of these methods is that they only consider greedy decision-making in 2D and structured indoor environments. 
In the domain of 3D aerial navigation, recent works have addressed target search by semantics-aware reconstruction of relevant objects \cite{papatheodorou_finding_2023} and efficient complete coverage of environments \cite{luo2024star} to find arbitrary targets. 
In contrast, we aim to leverage the MAV's capabilities for semantically-guided target search, as in the mentioned 2D methods, but for 3D unstructured environments.
While \cite{papatheodorou_finding_2023} uses semantic classes to determine the required reconstruction resolution of objects, we propose a planning pipeline that prioritizes viewpoints by propagating priorities of semantic classes into 3D space, similar to \cite{huang2022}. 
Our planning pipeline employs a combinatorial planner, generating efficient tours over frontier viewpoints similar to the TSP-based planner in \cite{luo2024star}. 
However, our weighted MLP formulation prioritizes semantically relevant frontiers, allowing for combining semantic target search and efficient exploration. 
In addition, we demonstrate the use of LLM-based similarity scores as a possible source of semantic priorities, as proposed by \cite{chen_how_2023}.

\subsection{Contribution}

The main contribution of this paper is a 3D semantic target search framework centered on a novel semantically-guided viewpoint planner designed to minimize expected search time.
The combinatorial planner prioritizes frontier viewpoints that likely lead to the target, as evaluated by an active perception pipeline that efficiently propagates semantic priorities from observed objects into continuous, occlusion-aware information gains of 3D candidate viewpoints.
Building on the FUEL exploration framework, our novel combinatorial planner formulation, which leverages both semantic and coverage viewpoint gains, generates semantically-guided exploration behavior.
We present proof-of-concept experiments in both simulation and real-world environments using a Micro Aerial Vehicle (MAV) that validate the effectiveness of our approach. 
Our quantitative results show that our method consistently outperforms exploration-only baselines in two distinct simulation environments, in terms of target search time and success rate.

\section{Preliminaries}

\subsection{Problem Formulation} \label{sec:prob_form}

An MAV is tasked with exploring a previously unseen 3D environment represented as a bounded volume $\cW \subset \R^3$ to find a target object in minimum time.
The MAV's pose in the environment at time instant $t$ 
is defined as $\textbf{x}_t \in SO(3)$, 
and we assume that fast and accurate position and attitude controllers are available, such that the robot can follow a trajectory by tracking pose increments \cite{mellinger2011minimum}.
It has a maximum linear velocity $v_{max}$, maximum acceleration $a_{max}$, and maximum yaw rate $\omega_{max}$. 
The robot is equipped with an RGB-D camera that provides a local observation of the environment.
At each time instance $t$ the robot receives a measurement tuple $z_t = (\textbf{x}_t, \mathcal{I}_c, \mathcal{I}_d$), where $\mathcal{I}_c$ and $\mathcal{I}_d$ are the RGB and depth images, respectively.

Using the RGB images, the robot can perform semantic segmentation to identify objects in the environment.
A set $\mathcal{S}$ of possible objects of interest (OOIs) represented by natural language semantic labels is available for segmentation.
Importantly, it is assumed that the objects in $\mathcal{S}$ have semantic relationships defined by a function $F: \mathcal{S} \times \mathcal{S} \rightarrow \mathbb{R}^+$, that quantifies how semantically related two objects in $\mathcal{S}$ are.
It can be exploited to guide the robot towards a target object $o^* \in \mathcal{S}$. A target is considered found when its relative semantic segmentation area in the robot's field of view crosses a threshold $\lambda_{min}$.

\noindent
\textbf{Problem Statement: }
Given a target object $o^*$, a bounded volume $\cW$, and the robot's initial pose $\textbf{x}_0$, find a collision-free global plan $\varrho$ through $\cW$ at each time instant t such that $o^*$ is discovered in minimum time, using the history of observations $z_{0:t}$ and the semantic relationship function $F$.

\subsection{Background}\label{sec:background}

\subsubsection{Frontier Exploration Planning} \label{sec:bkd_exploration}
The goal of robotic exploration is to efficiently build an occupancy map $\cM$ of a bounded volume $\cW$ using local range observations such as depth images.
This map is a 3D volumetric grid of voxels, with each voxel $m_k \in \mathcal{M}$ storing the probability of occupancy $P_k$. These probabilities are updated using an inverse camera sensor model, and Bayesian Inference \cite{han2019fiesta}.
Building an occupancy map of an unknown environment requires the robot to reduce the unexplored space in $\mathcal{M}$. 

Frontier-based exploration is an effective approach for reducing unknown space, which first detects a set $\mathcal{F}$ of \emph{frontiers}, i.e., boundary voxels between known and unknown space. Then, it directs the robot to observe these frontiers efficiently.
In this work, we build on the recent frontier-based MAV exploration method FUEL presented in \cite{zhou2021fuel}, which is introduced briefly below.

To maintain efficiency, frontiers are clustered into groups of voxels using a region-growing algorithm, leading to a set of clusters $\cK$, each with a minimum size $F_{min}$. That is, a cluster $K_i \in \cK$ with $K_i \subseteq \cF$ is only valid if $|K_i| \geq F_{min}$.
Around these clusters, a set $\mathcal{V}$ of \emph{viewpoints} is sampled, which are poses in free space that can 'view' the frontiers.
Viewpoints are further filtered using a minimum information gain $I_{min}$, considering only those that view at least $I_{min}$ frontier voxels.
Unknown space in $V$ can be reduced by finding the most efficient path through high-quality viewpoints. 
Recent frontier exploration methods such as FUEL \cite{zhou2021fuel}
and FAEL \cite{huang_fael_2023} use combinatorial optimization methods to plan a non-myopic global path through the viewpoints in $\mathcal{V}$.

\subsubsection{Semantic Relationships} \label{sec:bkd_semantic_priorities}
Semantics are labels or categories that humans use to classify objects. 
Humans use accumulated semantic knowledge to derive relationships between objects of interest when looking for targets. For instance, when searching for a \texttt{laptop}, we first look for a \texttt{table} as opposed to a \texttt{toilet} because the former is more correlated with the target object. 
These relationships are formalized by the semantic relationship function $F$
that maps a set $\mathcal{S}$ of semantic classes represented by language labels to scalar-valued similarity scores.

Large language models (LLM) or vision-language models like CLIP \cite{radford2021learning} or BERT \cite{devlin2018bert} 
can use their contextual understanding to infer such relationships.
These models use a neural network to first transform the text input to \emph{vector embeddings}, which are real-valued representations of the text in a high-dimensional feature space. 
Objects that often occur close to each other often have similar vector embeddings, as they appear in similar contexts in the training data.
Therefore, the cosine similarity score can be used to approximate the semantic relationship function $F$ between two labels,
obtained by calculating the dot product of their vector embeddings $\mathbf{a}$ and $\mathbf{b}$:
\begin{equation}
    \label{eq:cosine_sim}
    F (\mathbf{a}, \mathbf{b})
    = \dfrac{\mathbf{a} \cdot \mathbf{b}}{\|\mathbf{a}\|\|\mathbf{b}\|}
\end{equation}
Such similarity scores have been employed for target search in \cite{chen_how_2023,yokoyama_vlfm_2023}.


\section{Methodology}\label{sec:method}
\subsection{Overview}\label{sec:method_overview}

To motivate our method, consider how humans search for important objects. 
We infer target-object relations from the environment context, create a mental map of interesting objects, 
and then search \textit{near} these OOIs to find the target. 
For instance, we might search near a wardrobe to find clothes or under a table to locate a person during an earthquake (see \Cref{fig:priority_masking}).
Our method follows the same approach: To quickly find a target, it is essential to minimize unknown space \emph{near} objects of interest that are conceptually and spatially related to the target object. 
Therefore, we \emph{prioritize} search in specific regions, unlike coverage-based exploration, which tries to minimize \emph{all} unknown space. 
Additionally, balancing both tasks is necessary to ensure that the method works even under semantic uncertainty and does not incur significant costs in exploration time.
Our pipeline consists of three components. 
\begin{itemize}
    \item \textbf{Semantic Priority Masking}: This module processes the RGB image to segment objects, ranks them using a priority function, and compresses the segmentation image into a 2D priority mask (see \cref{sec:method_semantic_priorities} and \Cref{fig:priority_masking}).
    \item \textbf{Active Perception}: This module uses the generated priority mask, the depth image, and the drone's state to (a) sample
    a set of 3D frontier viewpoints in free space, and (b) evaluate the semantic information gain of each viewpoint (\cref{sec:method_mapping}). 
    \item \textbf{Target Search Planner}: This module solves a combinatorial optimization problem over the 3D viewpoints to create a global plan that prioritizes high-gain viewpoints, minimizing the expected search time (\cref{sec:method_WMLP}).
\end{itemize}
\Cref{fig:active_perception_pipeline} visualizes the active perception and planning parts of the pipeline.

\subsection{Semantic Priority Masking}
\label{sec:method_semantic_priorities}
The goal of this module is to use the RGB image $\mathcal{I}_c$ to generate a priority mask image $\mathcal{I}_p$ that has pixel-wise discrete priority values for each object of interest. 
A semantic segmentation image $\mathcal{I}_s$ and a set $\mathcal{S}_t \in \mathcal{S}$ of currently visible classes are generated using $\mathcal{I}_c$ at time $t$. 
The user-defined set $\mathcal{S}$ 
consists of a variety of relevant objects that can be encountered in the present scenario.
In this work, we assume the existence of a learning-based method, such as Mask-RCNN \cite{he2017mask} or Fast-SAM \cite{zhang2023faster}, that can generate $\mathcal{I}_s$ and $\mathcal{S}_t$. 

\begin{figure}
    \centering
    \includegraphics[width=\linewidth,trim={0 0 0cm 0cm},clip]{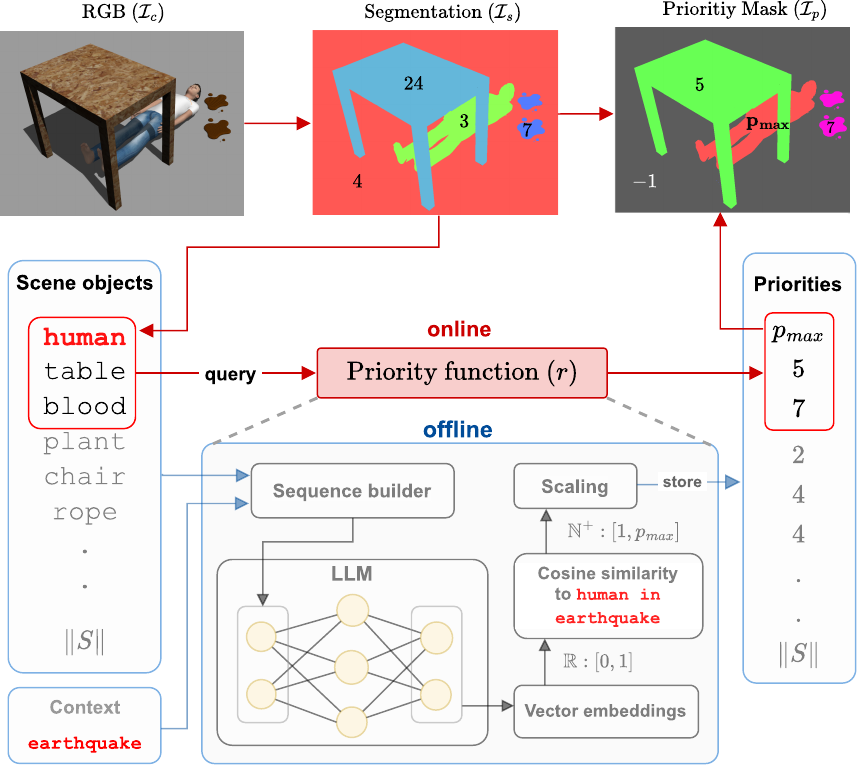}
    \caption{Semantic priority masking with LLM-based priority function. Red arrows and blue arrows represent online and offline operations, respectively. At runtime, the priority of each class in $\mathcal{S}_t$ is queried from a pre-computed priority vector to create the priority mask $\mathcal{I}_p$.}
    \label{fig:priority_masking}
\end{figure}

Instead of using $\mathcal{I}_s$ directly, we compress the semantic segmentation image into a priority mask $\mathcal{I}_p$. This image contains pixel-wise discrete integers for each class, indicating their relative importance to the target class. 
The process to generate this mask is demonstrated in \Cref{fig:priority_masking}. A priority mapping function $r:\mathcal{S}\rightarrow \mathbb{N}^+$ is derived \textit{offline}, for example, as detailed below, and then queried online.
The \textit{priority mask} $\mathcal{I}_p$ is created at runtime by replacing each pixel in $\mathcal{I}_s$ with its corresponding priority value.

\textbf{Offline Priority Inference: }
In the following, we describe a method to generate a semantic priority function $r$ using offline inference with an LLM.
First, situational context about the scenario is added to the target object label in the set $\mathcal{S}$,
using the formulation: [\texttt{label}] [\texttt{preposition}] [\texttt{context}], 
e.g., \texttt{human in earthquake}. This helps to derive more accurate relationships between the target object and other objects in the environment.
Second, each class in $\mathcal{S}$ is tokenized and passed through the LLM (bert-large-uncased \cite{devlin2018bert}) to produce an output tensor of embeddings, which is
averaged along the sequence dimension, resulting in $\boldsymbol{\tau}$ of size $\|\mathcal{S}\| \times n_e$. Here, each row represents a class in the embedding space. The target embedding vector is $\tau^*$.
To obtain cosine similarity scores for each class,
each vector in $\boldsymbol{\tau}$ is compared to $\tau^*$ using \cref{eq:cosine_sim}, producing values between 0 and 1.
Finally, the similarity scores are 
scaled to integer values within the range [1, $p_{\max}$]. 
Here, the maximum priority value $p_{\max}$ is a tunable parameter controlling the sensitivity of the semantic search.
The resulting priority mapping function $r$ from the set of classes to integer-valued priorities is 
stored offline, and then queried online with the set $\mathcal{S}_t$. 

\subsection{Active Perception}
\label{sec:method_mapping}
The goal of the active perception module is to use the priority mask $\mathcal{I}_p$, the depth image $\mathcal{I}_d$, and robot pose $\textbf{x}_t$ to create a set of viewpoints $\mathcal{V}$ in free space and corresponding information gains $I$ for each viewpoint. 

\begin{figure*}[t]
    \centering
    \includegraphics[width=\linewidth, trim=0 55 0 55, clip]{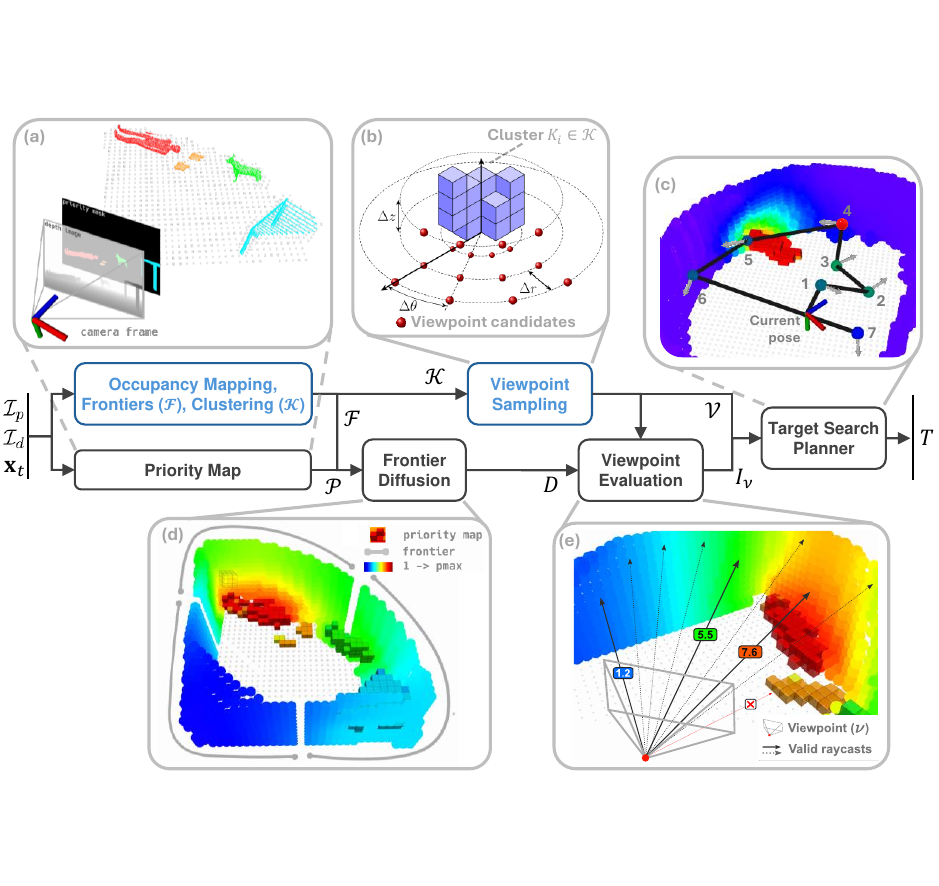}
    \caption{Active perception and planning pipeline. The blue blocks are from the FUEL framework \cite{zhou2021fuel}, and Subfigure (b) is inspired by \cite{zhou2021fuel}. In (d), the priority map voxels are cubes, and frontier voxels are spheres.}
    \label{fig:active_perception_pipeline}
\end{figure*}

Section \ref{sec:method_priority_map} provides a method for mapping of semantic priorities in 3D and
Section \ref{sec:method_frontier_diffusion} describes a method to diffuse semantic priorities to neighboring frontiers. 
Section \ref{sec:method_vpt_sampling} describes the process of generating viewpoints in free space, and Section \ref{sec:method_info_gain} describes our procedure to calculate information gain for each sampled viewpoint. 
\Cref{fig:active_perception_pipeline} shows an overview of the complete active perception pipeline.

\subsubsection{Priority Map}
\label{sec:method_priority_map}

The goal of this module is to represent the priority values in $\mathcal{I}_p$ in 3D space, where the drone will subsequently navigate and collect new observations.
The priority mask $\mathcal{I}_p$, the depth image $\mathcal{I}_d$, and robot pose $\textbf{x}_t$ are used to create a 4D position-intensity point cloud observation $\Omega$. 
Each point $\Omega_k = [\vq_w, p_w]^T$ in this point cloud is represented by the 3D position in the world frame $\vq_w = [x_w,y_w,z_w]^T$ and the priority value $p_w$ as the intensity channel. 
Let $\mathbf{d_i} = [u_i, v_i]^T$ be a pixel in the depth image $\mathcal{I}_d$ with depth value $z_i = \mathcal{I}_d(\mathbf{d_i})$ 
and $p_w = \mathcal{I}_p(\mathbf{d_i})$ be the corresponding priority value. 
The point cloud $\Omega$ is generated using standard projective transformations using the camera's intrinsic and extrinsic parameters.

The point cloud is also post-processed using voxel-grid filtering and statistical outlier removal. The priority value $p_w$ at each 3D point $k$ in $\Omega$ is then used to update a discrete volumetric grid $\mathcal{P}$ at the corresponding voxel $p_k \in \mathcal{P}$ using a simple weighted update (Eq. \ref{eq:priority_update})
\begin{equation}
     p_k \leftarrow (1-\alpha) p_k + \alpha p_w \label{eq:priority_update} \quad \forall k \in \{ 1, ..., |\Omega| \}.
\end{equation}
Here $\alpha$ is a learning rate that updates the map progressively and prevents noise from being integrated, and $|\Omega|$ is the size of the point cloud. 
Finally, a local section $\mathcal{P}_l \subset \mathcal{P}$ of the priority map centered around the drone is retrieved to keep the computational load bounded.

Representing semantics as a single-channel scalar priority avoids the computational complexity of high-dimensional 3D semantic mapping or instance tracking. This reduces the mapping task to a simple problem computationally comparable to standard occupancy mapping.

\subsubsection{Frontier Diffusion} 
\label{sec:method_frontier_diffusion}

The motivation for this section directly draws from the goal of minimizing unknown space \emph{near} objects of interest (Section \ref{sec:method_overview}). 
If frontiers can have increased importance near objects of interest, we can refine the search to interesting regions of the environment and find the target faster. 

This idea is implemented by diffusing the priority values from the local map section $\mathcal{P}_l$ into neighboring frontier voxels using a 3D partial convolution.
We use a partial convolution for the diffusion process as it normalizes over only the valid (non-empty) voxels in the sparse frontier structure \cite{liu2018image}.
A Gaussian kernel with spread $\sigma$, and size $W$ is used, thus making it a 3D Gaussian filter. 
\Cref{fig:active_perception_pipeline} shows a simulation example from RViz, where the diffusion process is applied to 3D frontiers.
The diffusion process is applied to each frontier voxel in a \emph{local} region surrounding the drone to maintain computational efficiency. 

Diffusion transforms discrete priority values into a continuous spatial utility function, preventing sparsely sampled viewpoints from missing important frontiers and being discarded by the planner (\cref{sec:method_WMLP}). Furthermore, diffusing priorities into the volumetric map enables semantic viewpoint evaluations to account for 3D occlusions via raycasting (\cref{sec:method_info_gain}), ensuring the planner prioritizes semantically relevant frontiers that are visually accessible.

The frontier diffusion module thus results in semantic priorities attached to each frontier voxel, that is, a frontier priority function 
$D: \mathcal{F} \rightarrow [1, p_{\max}]$,
which can further be used for downstream tasks like informative path planning.

\subsubsection{Viewpoint Sampling} 
\label{sec:method_vpt_sampling}
The goal of this section is to generate a set of viewpoints $\mathcal{V}$ which are candidate poses sampled in free space to 'view' the frontiers in $\cF$.
This module directly uses the approach from the FUEL framework \cite{zhou2021fuel}, which we briefly describe here. 
The viewpoint generation is based on the frontier clusters $\cK$ introduced in \cref{sec:bkd_exploration}.
Since even small regions of space can hold significance in semantic target search, we set the minimum cluster size threshold used in FUEL to $F_{\min} = 0$.
Frontier viewpoints $\nu \in \cV$ are generated by uniformly sampling free-space poses in a cylindrical coordinate system around the centroid of each frontier cluster (see \Cref{fig:active_perception_pipeline}c). 
The yaw angle of each viewpoint is set to maximize the sensor coverage of the frontier cluster. 
For more details, see \cite{zhou2021fuel}.

\subsubsection{Viewpoint Evaluation} 
\label{sec:method_info_gain}

This section describes a heuristic for computing a balanced coverage and semantic information gain for a viewpoint $\nu \in \mathcal{V}$, using the diffusion-based frontier priorities $D(f)$ of each frontier voxel $f \in \mathcal{F}$.

Consider \Cref{fig:active_perception_pipeline}d, where the frontier voxels are colored based on their priorities. 
Rays are cast from a candidate viewpoint $\nu$ toward the voxels in $\mathcal{F}$ to determine the priority value at the ends of valid rays.
A ray is considered valid when it is unobstructed by occupied or unknown space in $\cM$.
Voxels at the end of valid rays create a new \emph{visible} frontier set $\mathcal{F}_\nu \subset \mathcal{F}$ for each viewpoint $\nu \in \cV$. 
The frontier priority value $D(f)$ for each $f \in \mathcal{F}_\nu$ is then passed through a transfer function $\Phi$  and summed up to give the total information gain $I(\nu)$ of the viewpoint $\nu$, as in
\begin{align}
    \Phi(f) &= \max{\Bigl\{
        \exp{\bigl(
            \gamma(D(f)-1)    
        \bigr)}, 1
    \Bigr\}} 
    \label{eq:info_transfer} 
    \\
    I(\nu) &= \sum_{f\in \mathcal{F}_\nu} \Phi(f),
    \label{eq:info_gain}
\end{align}
where $\gamma \in \mathbb{R}^+$ is a tuning parameter.

To demonstrate this, 
consider high-priority frontier voxels near semantic objects (as in \Cref{fig:active_perception_pipeline}d), which are exponentially weighted by the transfer function $\Phi$.
Thus, viewpoints oriented towards semantically meaningful regions of the 3D space achieve a high information gain $I_\nu$ and can be prioritized in semantic target search. 
Conversely, consider a frontier $f$ far from semantically interesting objects with the lowest priority value, i.e., $D(f)=1$, resulting in a volumetric coverage gain $\Phi(f)=1$.
The parameter $\gamma$ weights the semantic priority against the coverage gain: 
a higher $\gamma$ increases the difference between semantically relevant and irrelevant areas, while a lower $\gamma$ shifts the objective towards coverage exploration. For $\gamma=0$, the viewpoint gain equals the number of covered voxels, i.e., $I(\nu)=|\mathcal{F}_\nu|$.

The described method for information gain evaluation can integrate semantic priorities with volumetric coverage and uses a tunable exponential weighting function to ensure that semantically relevant viewpoints are prioritized.

\subsection{Target Search Planner} \label{sec:method_WMLP}

The goal of the global target search planner is to use the set of viewpoints $\mathcal{V}$ and their respective information gains $I$ generated in the active perception module together with the drone's state $\textbf{x}_t$ to plan a global path that minimizes the time to find the target object.

Our approach extends the idea of combinatorial planning between different viewpoints \cite{zhou2021fuel,huang_fael_2023} to determine their optimal visitation order. While FUEL \cite{zhou2021fuel} uses a TSP  that minimizes the total traveling distance, semantic target search needs to prioritize viewpoints with high semantic information gain, which are expected to be close to the target.
To this end, we propose to formulate the combinatorial target search problem as a variant of the Minimum Latency Problem (MLP) \cite{blumMinimumLatencyProblem1994} that minimizes the average waiting time, or
\textit{latency}, of multiple tasks, which in our case are the viewpoints. 
Specifically, our \textit{weighted} MLP (WMLP) formulation prioritizes minimizing the latency of semantically promising viewpoints, using the information gains $I(\nu)$ as weights for each viewpoint $\nu \in \mathcal{V}$. 
This planner formulation is similar to our concurrent work \cite{lodel_priorities_2026}, where we focus on leveraging priorities learned from human inputs, while this paper emphasizes efficient target search with an MAV.

The WMLP is formulated over a modified set of viewpoints $\mathcal{V}^*$:
Firstly, we only consider viewpoints with a minimum information gain $I_{\min}$, i.e., $\mathcal{V}' = \{\nu \in \mathcal{V} \mid I(\nu) \geq I_{\min}\}$, similar as done in \cite{zhou2021fuel}.
Secondly, we add the drone's current pose $\textbf{x}_t$, i.e., $\mathcal{V}^* = \mathcal{V}' \cup \{ \textbf{x}_t \}$,
planning over $N = |\mathcal{V}^*|$ poses.
The tour $T$ describes a visitation order of the viewpoints $\nu \in \mathcal{V}^*$, 
where $T(i) = \nu_j$ denotes the $i^{th}$ viewpoint in the tour being $\nu_j$, for $i,j \in \{0,\cdots,N-1\}$.
By definition, $T(0) = \vx_t$,
as each tour starts at the drone's current pose.
Let $\mathbf{C} \in \R^{N \times N}$ be a cost matrix quantifying the traveling time between viewpoints, where $C_{ij}$ is the time required to move from viewpoint $\nu_i$ to $\nu_j$.
Then the WMLP is formulated as
\begin{align}
    \min_T\
    \sum_{i=1}^{N}
    I(T(i))
    \sum_{j=1}^{i}
    C_{T(j-1),T(j)},
    \label{eq:mlp_cost}
\end{align}
where the inner sum computes the latency until visiting the $i^{th}$ viewpoint $T(i)$,
and the tour cost function is a weighted sum of these latencies using the viewpoints' information gains $I(T(i))$. This formulation extends the classical MLP with uniform weights to weighted task latencies.

Solving the problem in \cref{eq:mlp_cost} means that viewpoints $\nu$ with higher information gain $I(\nu)$ get scheduled earlier in the tour. 
Since the information gain was calculated by balancing coverage and semantic priorities in \cref{sec:method_info_gain}, this objective pursues both coverage and target search, and is thus more robust to semantic uncertainty.

While defining the cost matrix $\mathbf{C}$ using Euclidean distances is common in TSP formulations and robotic exploration \cite{huang_fael_2023,cao2023tare}, it does not account for the complex dynamics of MAVs.
Thus, we use the kinematic cost function from \cite{zhou2021fuel} to compute the traveling times in the cost matrix $\mathbf{C}$.
$C_{ij}$ is then defined as the minimum time required to switch between two viewpoints $\nu_i, \nu_j \in \mathcal{V}^*$. 
\begin{equation}
    C_{ij} = \max \left(\dfrac{\text{length}(\nu^{\textbf{p}}_i, \nu^{\textbf{p}}_j)}{v_{\max}}, \\\dfrac{|\nu^{\psi}_i - \nu^\psi_j|}{\omega_{\max}}\right) \label{eq:cost_mat}
\end{equation}
Here, $\nu^{\textbf{p}}_i$ is the 3D position, $\text{length}(\nu^{\textbf{p}}_i, \nu^{\textbf{p}}_j)$ computes the path length between $\nu^{\textbf{p}}_i$ and $\nu^{\textbf{p}}_j$,
and $\nu^{\psi}_i$ is the yaw angle of the $i^{th}$ viewpoint from set $\mathcal{V}^*$.

The objective in \cref{eq:mlp_cost} is approximately solved using a large neighborhood search (LNS) algorithm \cite{pisingerLargeNeighborhoodSearch2019} that iteratively destroys and reconstructs the tour $T$.
In each iteration, the custom LNS algorithm randomly removes up to 30\% of the tour and then reconstructs it using the
cheapest insertion heuristic \cite{rosenkrantz1977} followed by a 2-opt swapping search \cite{croesMethodSolvingTravelingSalesman1958}.

In summary, the tour $T$ minimizes the time to arrive at 
regions of the environment that are semantically important with respect to the target object, thus approximately minimizing target search time.
\section{Experimental Setup}
\label{sec:setup}

\subsection{Simulation Environments} \label{sec:setup_simulation}
\begin{figure}[t]
    \centering
    \includegraphics[width=\linewidth]{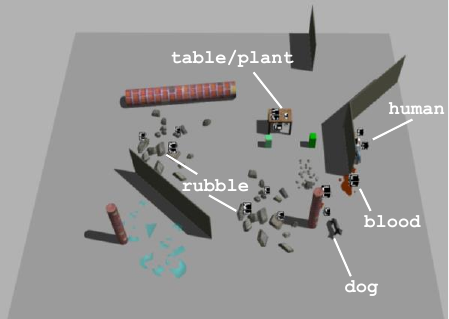}
    \caption{Earthquake environment (Gazebo).}
    \label{fig:earthquake_gazebo}
\end{figure}

\begin{figure}
    \centering
    \includegraphics[width=\linewidth]{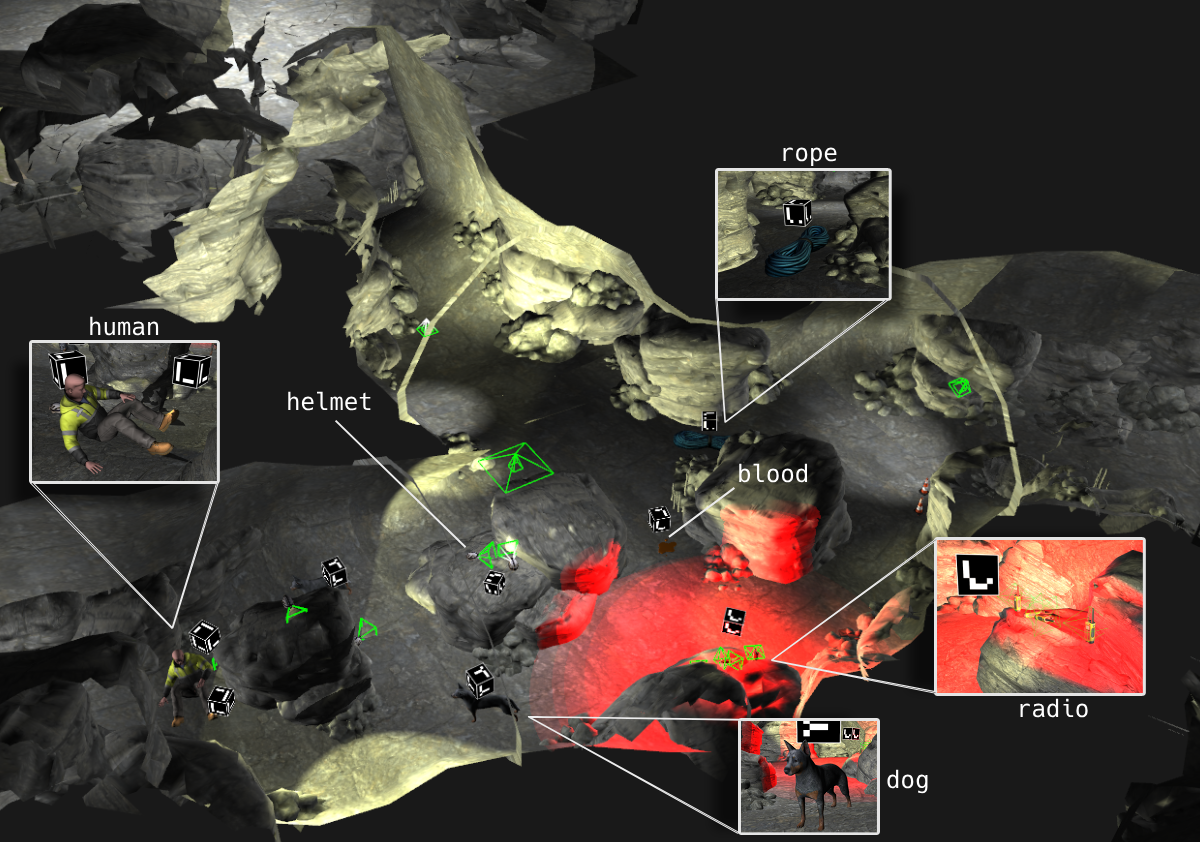}
    \caption{Cave environment (Gazebo).}
    \label{fig:cave_gazebo}
\end{figure}

Two realistic simulation environments were used to evaluate the algorithm in the PX4-Gazebo SITL simulator\footnote{\url{docs.px4.io/main/en/simulation/ros_interface.html}}. 
The \emph{Earthquake} is a custom environment (\Cref{fig:earthquake_gazebo}), and the \emph{Cave} (\Cref{fig:cave_gazebo}) is a section of the 'Cave Circuit 02' world from the DARPA SubT Challenge\footnote{\url{https://www.darpa.mil/program/darpa-subterranean-challenge}}.

Both environments contain a trapped human as the target, sufficiently occluded to make the problem challenging.
A common superset of objects was used as semantic clues for both environments. 
This set combines object classes from the DARPA SubT challenge and common sense objects that are expected near a \texttt{human} target in search and rescue situations. These objects are placed in the environment according to their semantic relationship with the target, e.g., blood is the closest to the human, while the rubble is further away.
We also included unimportant distractor objects like \texttt{toy} and \texttt{plant}
to evaluate whether the planner prioritizes semantically relevant objects rather than any observed object.

We used ArUco markers as a robust proxy for 2D semantic segmentation. This decouples the evaluation of our target search pipeline from the performance fluctuations of specific segmentation models,
ensuring that the results reflect the efficiency of the search strategy rather than the robustness of the vision module.
These markers were placed near their respective semantic objects, and the 2D segmentation image then contains a pixel-wise label for each marker $o \in \mathcal{S}$. 

Both environments were made sufficiently large to ensure realistic exploration,
and the start pose was chosen to be far from the target to observe the effect of semantic exploration.  
Since we evaluate the experiments on the absolute time-to-target metric rather than a relative metric (see \cref{sec:results_metrics}), starting from random start poses in a relatively small environment will not give comparable results. 
However, it was observed that there is substantial variance in the SITL simulation (\cref{sec:setup_arch}), primarily caused by non-deterministic communication between different software nodes, resulting in different observed trajectories for multiple runs with the same start pose.
Therefore, we ran the same experiment multiple times for quantitative evaluation in each environment instead of varying the start pose.
In the simulation experiments, we use the offline LLM inference described in \cref{sec:method_semantic_priorities} as the source of the semantic priority function $r$.

\subsection{Software Architecture}\label{sec:setup_arch}

\begin{figure}[t]
    \centering
    \includegraphics[width=\linewidth]{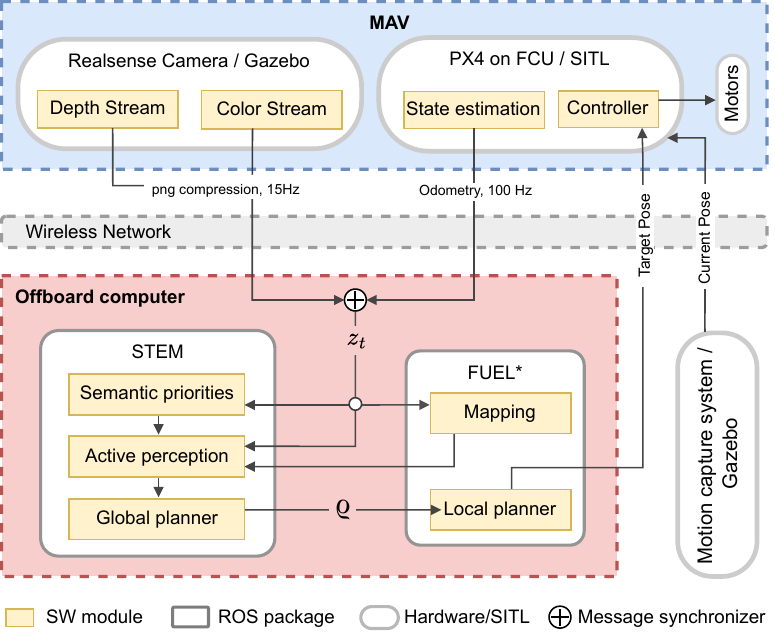}
    \caption{Software architecture. A modified version of FUEL \cite{zhou2021fuel} is used for mapping and local planning.}
    \label{fig:sw_arch}
\end{figure}

Our software stack is based on the Robotics Operating System (ROS) and integrates our target search method described in \cref{sec:method} with the modified FUEL exploration pipeline \cite{zhou2021fuel}.
The architecture is demonstrated in \Cref{fig:sw_arch}. 
The FUEL pipeline contributes the implementations for 3D voxel mapping, frontier detection and clustering, and viewpoint sampling (see also \Cref{fig:active_perception_pipeline}), as well as the local planner.
A key capability of the software is that we use the same pipeline for both hardware and software experiments, with the only difference being the source of the measurement tuple $z_t$. 
For simulation, this measurement comes from the Gazebo simulator, whereas for hardware experiments, this measurement comes from the onboard camera and 
state-estimation based on motion capture and IMU information. 
The parameters used by the planning pipeline are summarized in \cref{tab:setup_params}.

\begin{table}[tbp]
\vspace{-1.0em}
    \centering
    \caption{Parameter values for our pipeline used in all experiments.}
    \label{tab:setup_params}
    \small
    \begin{tabular}{ll|ll}
         \toprule
         Parameter & Value & Parameter & Value \\
         \midrule
        $\alpha$ & 0.9 &  
        $v_{max}$ & 0.5 m/s \\ 
        $a_{max}$  & 0.5 $m/s^2$ &  
        $\omega_{max}$ & 0.7 rad/s \\
        $h$ & 480 &
        $w$ & 848 \\  
        $\sigma$ & 2 &  
        $W$ & 5 \\  
        $p_{max}$ & 8 &  
        $\gamma$ & 4 \\  
        $I_{min}$ & 10 &  
        $\lambda_{min}$ & 0.01 \\ 
        $\| \mathcal{S} \|$ & 22 & 
        $F_{min}$ & 0 \\
         \bottomrule
    \end{tabular}
\end{table}

\subsection{Hardware Setup}\label{sec:setup_rw}
\begin{figure}[t]
    \centering
    \includegraphics[width=\linewidth]{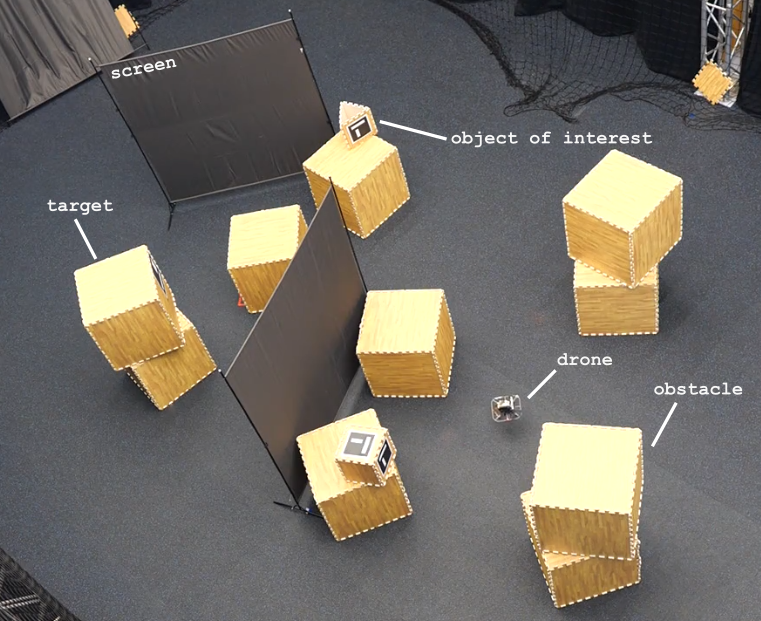}
    \caption{Lab environment for hardware experiments}
    \label{fig:lab_setup}
\end{figure}

Hardware experiments were performed with a custom-built Micro Aerial Vehicle (MAV) previously used in \cite{chen2023_risk_aware} and modeled for the Gazebo simulation experiments. The MAV is equipped with an Intel RealSense D455 camera and an Nvidia Jetson Xavier NX onboard computer. A HolyBro Kakute F7 V2 flight controller was used with PX4 autopilot software. The MAV was localized in the environment via a Vicon motion capture system. ArUco markers were placed in the environment as semantic objects of interest, and the experiments were conducted in sufficiently cluttered configurations with screens and boxes as obstacles. 
The lab environments are created to be geometrically challenging due to obstacles and occlusions, but do not represent real scenarios with actual semantic relationships.
Therefore, in the hardware experiments, we do not use LLM-based priority inference (see \cref{sec:method_priority_map}) used in the simulation experiments, but instead provide a handcrafted priority function $r$.
Note that in this case, the whole pipeline, including priority masking, active perception, and planning, is still validated in the hardware experiments
\Cref{fig:lab_setup} shows an exemplary environment for the experiment.

\section{Simulation Results}
\label{sec:results}

In this section, we present and discuss the results of our proposed method \textit{STEM} on the simulation environments from \cref{sec:setup_simulation}. 
Evaluation metrics are proposed in \cref{sec:results_metrics}, and two baselines are compared in \cref{sec:results_baselines}. 
We provide the primary performance comparisons with baselines in \cref{sec:results_quant}, which are further supported using qualitative results in \cref{sec:results_qual}. 
Finally, we also conduct an ablation study to analyze the importance of the proposed target search planner in \cref{sec:results_ablations}.

\subsection{Evaluation Metrics}\label{sec:results_metrics}
We employ the following metrics: 

\textbf{Success \%}:
This metric calculates the percentage of episodes in which the target was successfully found across $n$ trials. 
An object $o$ is considered found when the fraction of object pixels $\lambda_o$ crosses a threshold $\lambda_{min}$ in the segmentation image, with $\lambda_o$ defined as
$\lambda_o = \beta / (wh)$.
Here, $\beta$ is the number of pixels belonging to the object's ArUco marker, and $w$ and $h$ are the segmentation image width and height, respectively. The parameter $\lambda_{min}$ depends on the environment complexity and camera intrinsics. 
For example, a value of $\lambda_o=0.02$ means that the marker occupies 2\% of the field of view in the image plane.

\textbf{Time to target}:
A commonly used metric for ObjectNav tasks is Success weighted by Path Length (SPL) \cite{batra_objectnav_2020}. A notable drawback of this metric is that it only considers travel distance in SE(3), and for robots with complex dynamics (such as MAVs), the completion time is recommended \cite{yokoyama2021success}. 
Thus, we record the first time instant when a target $o^*$ was successfully detected (i.e., $\lambda_{o^*} \geq \lambda_{min}$) and call this metric the Time to target $t^*$.

\textbf{Exploration time}:
Since balancing exploration and target search is a secondary goal for our method, we also measure the exploration time $t_f$ in seconds. An environment is considered explored when no \emph{visible} frontier can be found for 10 consecutive iterations. For a frontier to be considered \emph{visible}, it must have (1) at least $F_{min}$ number of clustered voxels, and (2) at least one viewpoint with minimum information gain $I_{min}$. These conditions are also used by the authors of FUEL
\cite{zhou2021fuel}. \looseness=-1

\subsection{Baselines}\label{sec:results_baselines}
We evaluate our method by comparing it with the highly efficient coverage-based exploration pipeline FUEL \cite{zhou2021fuel}, since there are no semantically guided target search methods for MAVs. This allows us to isolate the performance gain achieved by adding semantic guidance to a coverage framework.
Our method is compared to two versions of the FUEL pipeline \cite{zhou2021fuel}, 
differing by the two parameters $F_{min}$ and $I_{min}$:
The \textit{FUEL-original} baseline uses the original parametrization with 
$F_{min} = 100$ and $I_{min} = 20$ as proposed in \cite{zhou2021fuel}.
For the \textit{FUEL-complete} baseline, these parameters were tuned to maximize target search success, as we noticed that finding the target in small regions depended strongly on these parameters. For the Earthquake scenario, $F_{min} = 0$ and $I_{min} = 5$ were used, and for the Cave scenario, $F_{min} = 0$ and $I_{min} = 0$. Additionally, frontier down-sampling (see \cite{zhou2021fuel}) was turned off due to the narrow passages and small frontier sizes in the Cave environment.
The sensor range $R_{max}$ for both baselines was kept the same as our method to make comparisons fair.

\subsection{Performance Results} \label{sec:results_quant}
This section presents the performance results of our method compared to the two baselines introduced in  \cref{sec:results_baselines} using the metrics from \cref{sec:results_metrics}. Table \ref{tab:comparison} summarises results for both the Earthquake and Cave simulation environments. The data was gathered over 20 trials for each method and environment. 
The results are discussed below.

\begin{table*}[t]
    \caption{
        Comparison study with baselines in the Earthquake and Cave environments. Two baselines from \cref{sec:results_baselines} were compared to our method on success \%, time to target $t*$, and exploration time $t_f$. 
    }
    \label{tab:comparison}
    \centering
    \begin{tabular}{c|l|cc|c}
    \toprule
    \multicolumn{2}{c}{} & \multicolumn{2}{|c|}{Target search}&  Exploration\\
    \midrule
     Env & Method & Success \% & Time $t^*$ (s)  & Time $t_f$ (s)\\
     \midrule    
     \multirow{4}{*}{\makecell{Earthquake\\ \\}}    
         & FUEL-original & 5\%  &   $79.8 \pm 0.0$  & $\bf{107.3 \pm 17.3}$ \\
         & FUEL-complete & 100\%  &   $76.6 \pm 33.5$  & $129.4 \pm 12.9
         $  \\
         & STEM (Ours) & \textbf{100\%}  &  $\bf{56.9 \pm 22.8} $ & $139.9 \pm 11.9$\\ 
     \midrule
      \multirow{4}{*}{\makecell{Cave\\ \\}}    
         & FUEL-original & 40\% &  $65.0 \pm 14.8$  &   $\bf{93.6 \pm 30.7}$ \\    
         & FUEL-complete & 90\% &   $95.3 \pm 26.6$  & $166.4 \pm 31.5$ \\
         & STEM (Ours) &\textbf{90\%}  & $\bf{64.1 \pm 20.3}$ & $130.7 \pm 19.8$ \\
     \bottomrule
    \end{tabular}
\end{table*}

STEM consistently finds the target faster than all methods and is as successful as the FUEL-complete baseline, which is tuned for high success rates.
Moreover, it keeps the exploration times within reasonable bounds. 
The fast target search times and high success rates have been achieved by the combination of the semantic viewpoint evaluation and the combined target search planner.
The viewpoint evaluation
based on the semantic frontier priorities diffused from relevant objects 
and subsequent filtering of viewpoints using $I_{min}$ leads to a viewpoint set biased towards viewing regions of high likelihood of target presence.
This allows our method to find the target consistently.
The combinatorial target search planner schedules high semantic gain viewpoints earlier, such that the MAV quickly reaches regions where the target is likely to be found.
Conversely, FUEL-complete does not use semantic information to guide search and explores irrelevant regions first, therefore taking more time to find the target while still achieving high success rates.

The performance difference between FUEL-complete and STEM 
underlines the importance of the viewpoint gain threshold $I_{min}$:
A lower $I_{min}$ retains viewpoints for small frontier clusters in tight spaces, 
and therefore enabling a high success rate for FUEL-complete.
However, this also means that many small but unimportant regions are covered, leading to inefficient and slow target search and exploration, as reflected by the exploration time results of FUEL-complete.
This emphasizes the advantage of evaluating the semantic relevance of frontiers, as it allows retaining and prioritizing viewpoints in small but important regions while ignoring small, unimportant regions.

FUEL-original rarely finds the target 
due to its increased minimum frontier size $F_{min}$ and viewpoint threshold $I_{min}$. As a result, frontier clusters or viewpoints in tight spaces that lead to the target are ignored.
However, this baseline consistently completes exploration faster,
since the clustering and thresholding facilitate more stable viewpoint sets $\cV$ and more consistent global plans, allowing the MAV to maintain high speeds throughout the episode.
Furthermore, FUEL-original solves a metric TSP using the LKH heuristic solver, generating robust and efficient global plans.

\subsection{Qualitative Results} \label{sec:results_qual}
In this section, we visualize the behavior of our proposed method compared to the baselines to support the quantitative results from \cref{sec:results_quant}. 

\begin{figure*}[t]
    \centering
    \begin{subfigure}[b]{0.329\linewidth}
        \centering
        \includegraphics[width=\textwidth, trim=250 100 500 150, clip]{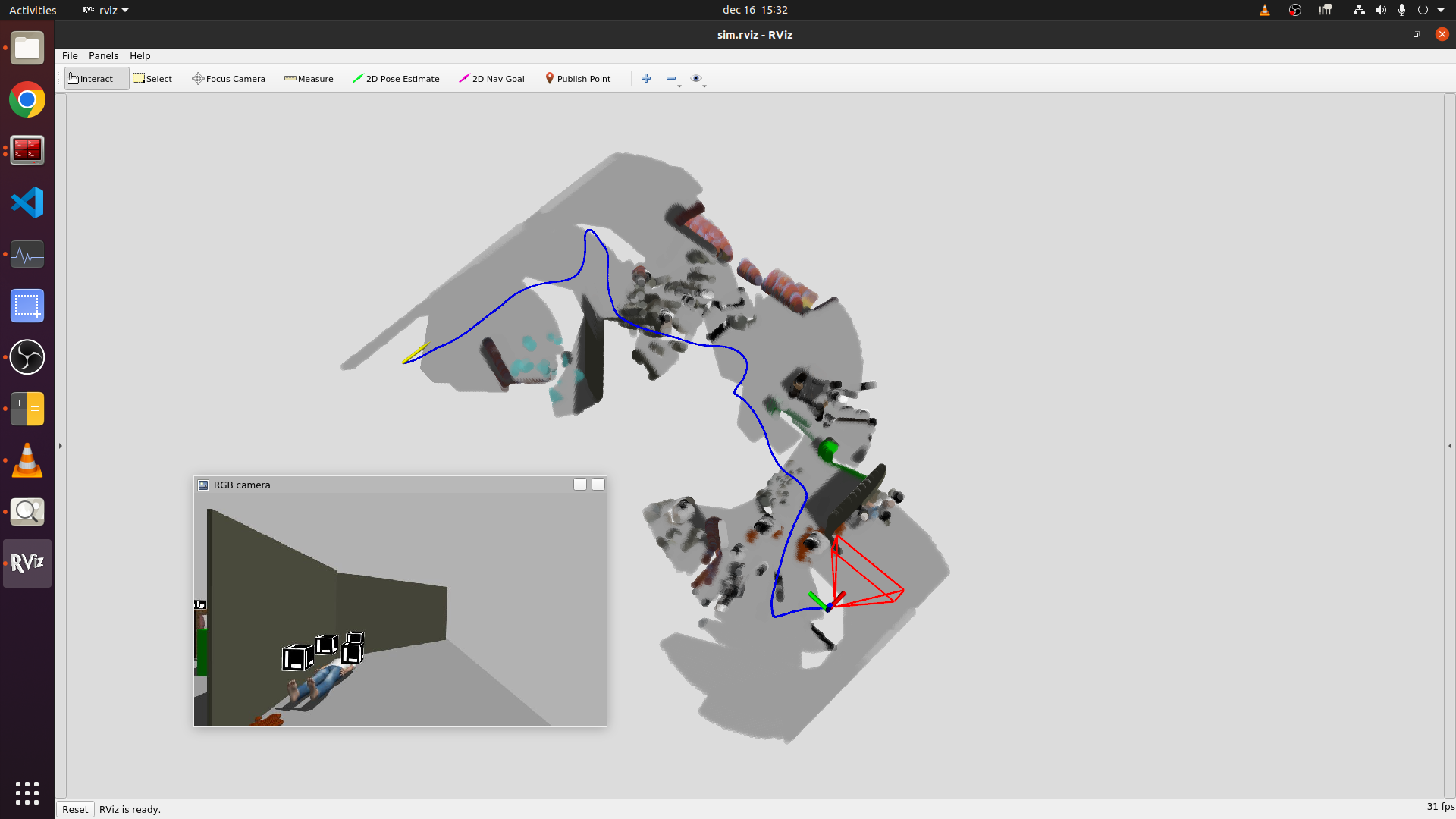}
        \caption{STEM}
        \label{fig:eq_ours_episode}
    \end{subfigure}
    \begin{subfigure}[b]{0.329\linewidth}
        \centering
        \includegraphics[width=\textwidth, trim=250 100 500 150, clip]{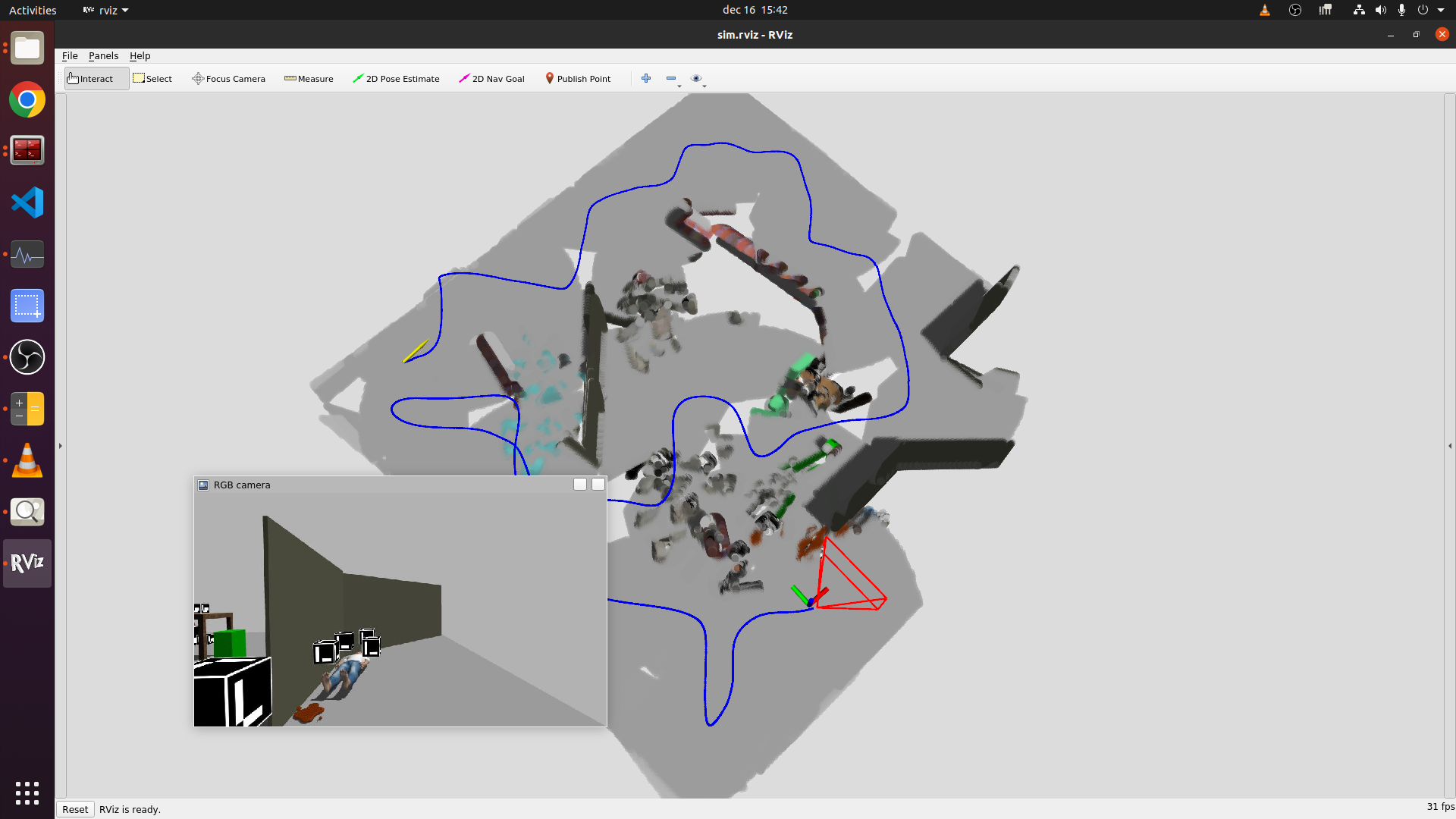}
        \caption{FUEL-complete}
        \label{fig:eq_fuel_mod_episode}
    \end{subfigure}
    \begin{subfigure}[b]{0.329\linewidth}
        \centering
        \includegraphics[width=\textwidth, trim=250 100 500 150, clip]{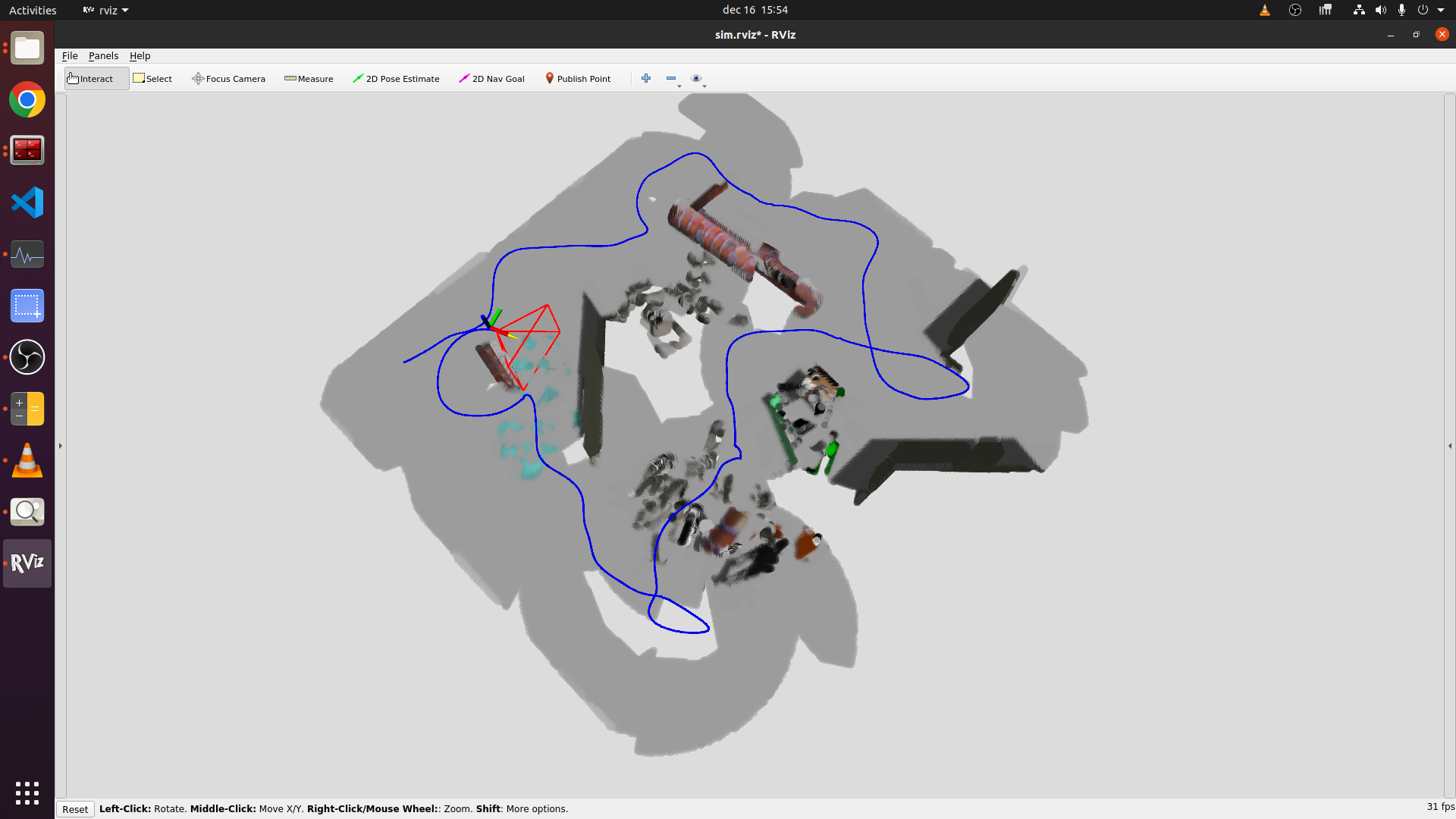}
        \caption{FUEL-original}
        \label{fig:eq_fuel_episode}
    \end{subfigure}
    \caption{Qualitative comparison with baselines for the Earthquake environment. Episodes were recorded until $t^*$ or $t_f$, whichever comes first. The reconstructed point cloud from RViz is shown along with the drone's trajectory (in blue). The camera FOV is shown in red, and the latest RGB camera image is displayed.}
    \smallskip
    \small
    \label{fig:eq_comparison_episodes}
\end{figure*}

\begin{figure*}[t]
    \centering
    \includegraphics{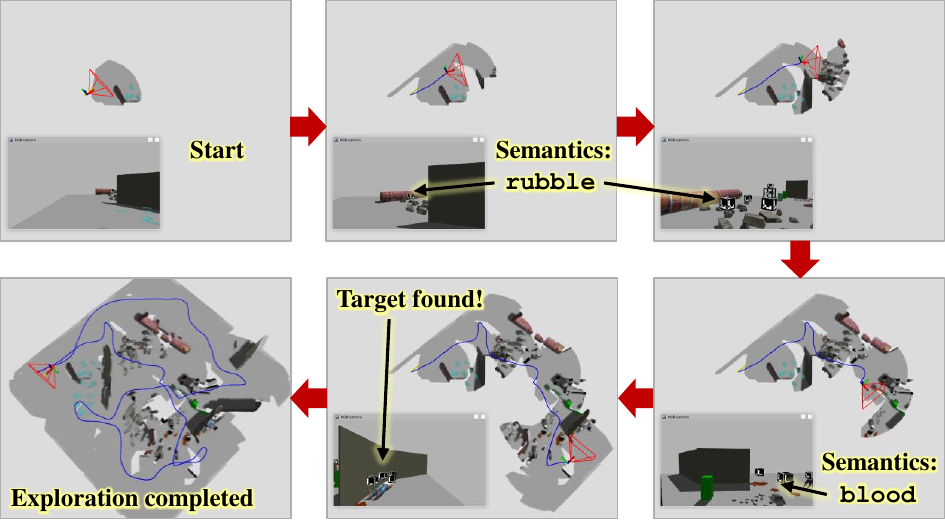}
    \caption{
    Key frames of target search using STEM in the Earthquake environment. 
    The snapshots show the recorded RGB point cloud, the camera FOV in red, the trajectory in blue, and the current RGB image.
    }
    \label{fig:earthquake_keyframes}
\end{figure*}

\begin{figure*}[t]
    \centering
    \begin{subfigure}[b]{0.329\linewidth}
        \centering
        \includegraphics[width=\textwidth, trim=400 20 440 160, clip]{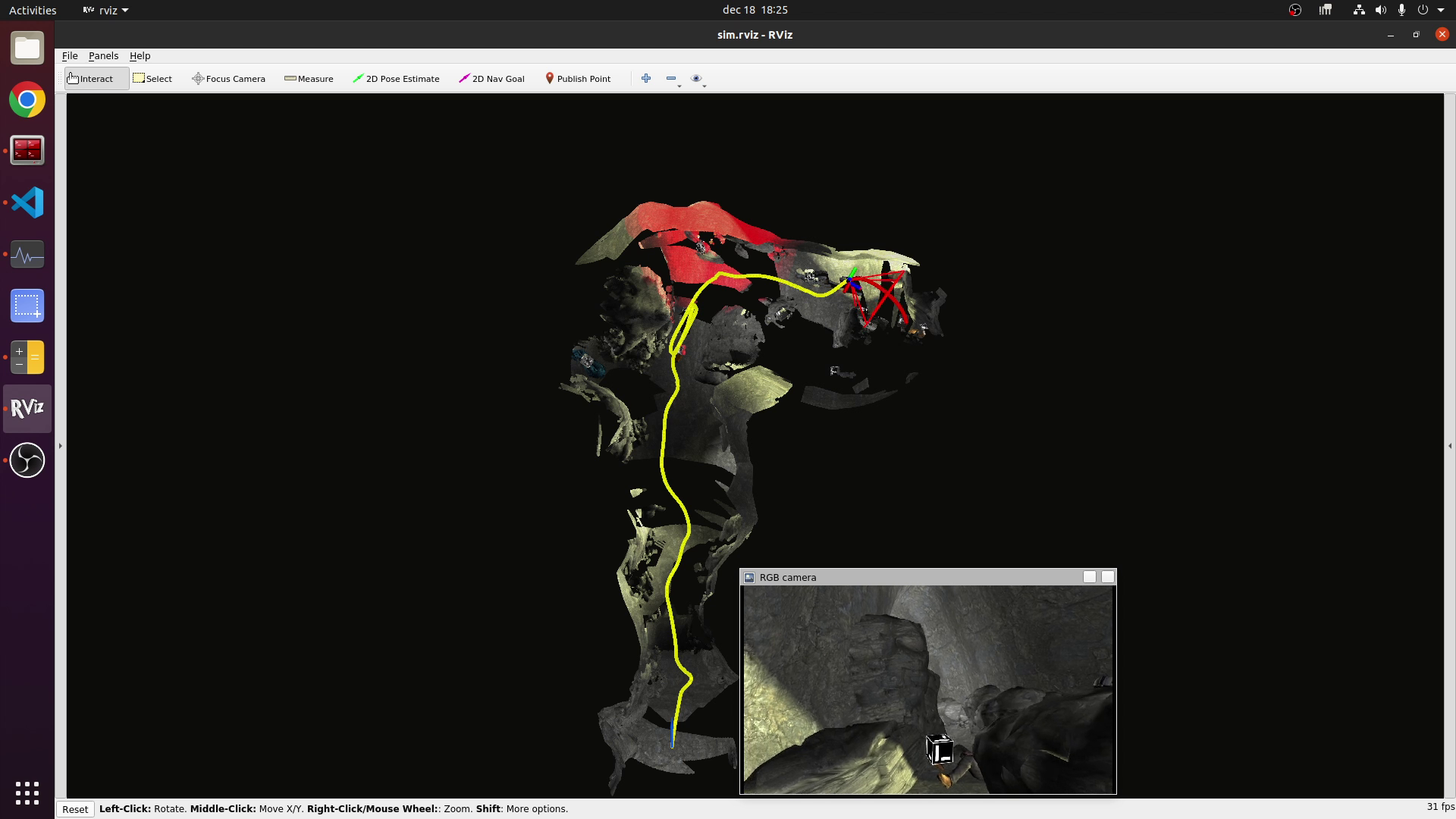}
        \caption{STEM}
        \label{fig:cave_ours_episode}
    \end{subfigure}
    \begin{subfigure}[b]{0.329\linewidth}
        \centering
        \includegraphics[width=\textwidth, trim=400 20 440 160, clip]{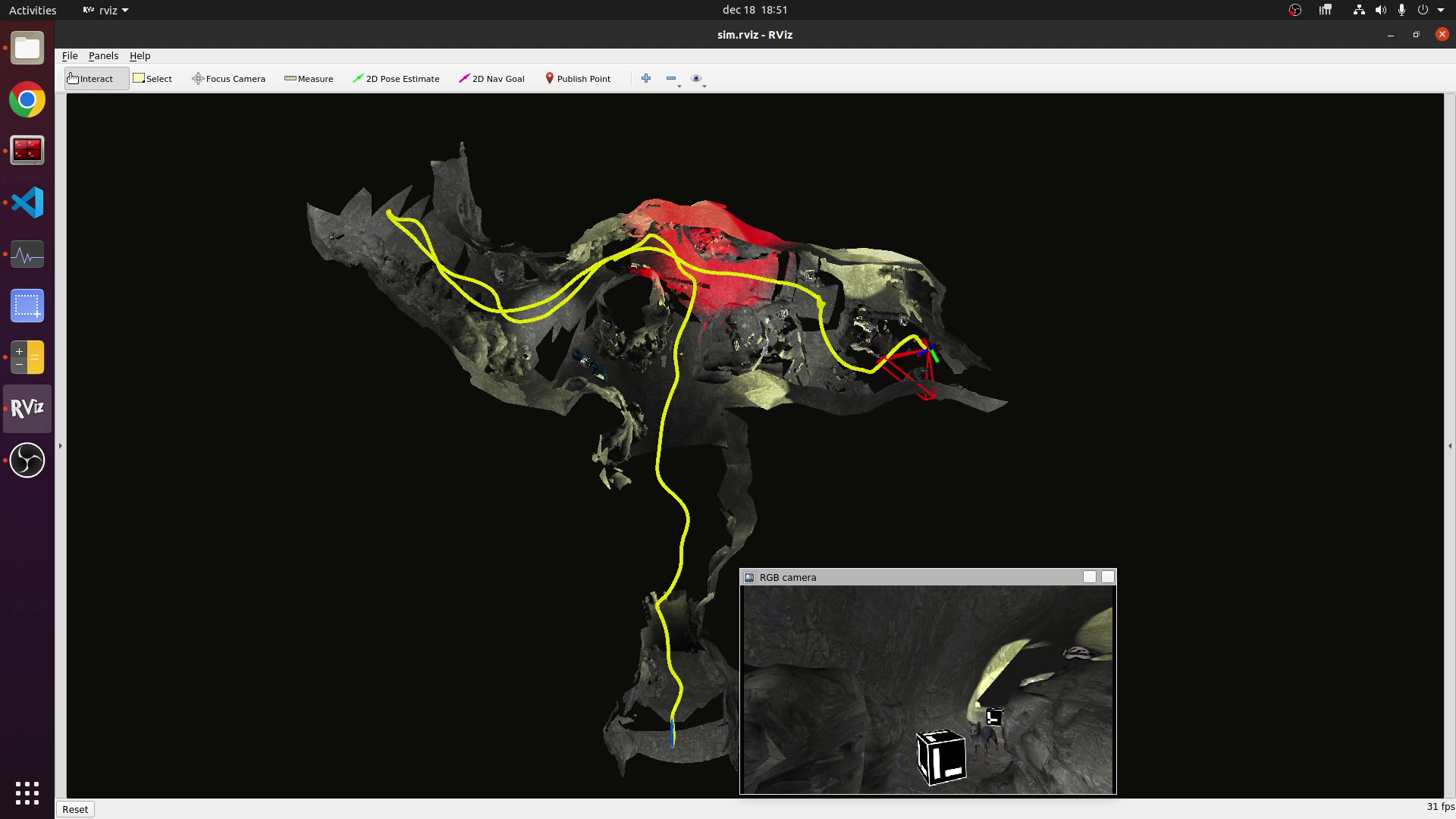}
        \caption{FUEL-complete}
        \label{fig:cave_fuel_mod_episode}
    \end{subfigure}
    \begin{subfigure}[b]{0.329\linewidth}
        \centering
        \includegraphics[width=\textwidth, trim=400 20 440 160, clip]{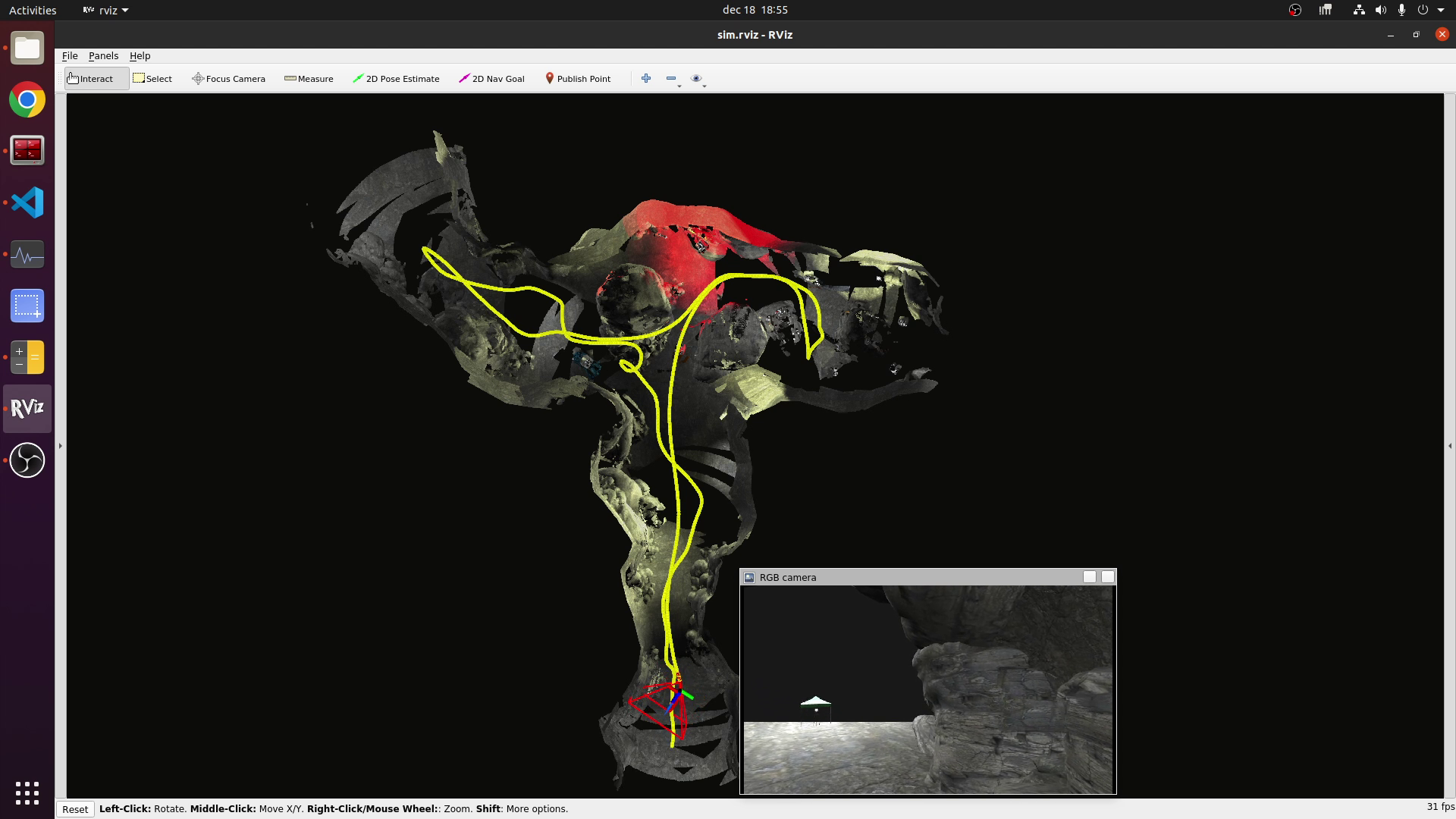}
        \caption{FUEL-original}
        \label{fig:cave_fuel_episode}
    \end{subfigure}
    \caption{Qualitative comparison with baselines for the cave environment. Episodes were recorded until $t^*$ or $t_f$, whichever comes first. The reconstructed point cloud from RViz is shown along with the drone's trajectory (in yellow). The camera FOV is shown in red, and the latest RGB camera image is displayed.}
    \smallskip
    \small
    \label{fig:cave_comparison}
\end{figure*}

\Cref{fig:eq_comparison_episodes} shows a qualitative comparison of the trajectories produced by the three methods in the Earthquake environment until the target is found or exploration is completed.
The results show that STEM finds the target quickly without taking large detours, while FUEL-complete explores large parts of the environment before finding the target, and FUEL-original does not find the target.
STEM achieves these results by focusing exploration on regions semantically related to the target, while both FUEL methods only aim at maximizing coverage gains.
The trajectory of FUEL-complete shows that it came close to the target and highly relevant semantic cues (\texttt{blood} objects), but does not react to these cues and continues exploring irrelevant regions.

Comparing FUEL-complete and FUEL-original emphasizes the effect of the thresholds $F_{min}$ and $I_{min}$ discussed in \cref{sec:results_quant}. 
When FUEL-original does not find any sufficiently large frontiers or viewpoints and completes exploration, there are still several small regions of the map left unexplored. FUEL-complete, however, is more thorough in its exploration but explores the environment less efficiently, as evidenced by more directional changes in its trajectory.

Figure \ref{fig:earthquake_keyframes} further shows keyframes of our algorithm performing target search in the earthquake environment.
The MAV starts at a disadvantaged position and starts to explore first to gather information. 
When it comes across semantically relevant objects such as \texttt{rubble} or \texttt{blood}, it samples informative viewpoints covering the nearby frontiers. 
Planning a path through such viewpoints, continually using the target search planner, allows the MAV to find the hidden target quickly. 
In addition, the keyframes show that our approach correctly ignores the distractor objects like "plant".

Figure \ref{fig:cave_comparison} presents the trajectories of our method and the two baselines in the cave environment. 
The results show that STEM is able to handle a complex 3D environment with many occlusions and tight spaces, which underlines its ability to balance exploration and semantic target search.
After exploring initially, the MAV comes across objects such as \texttt{radio} and \texttt{dog} in the right arm of the cave, which STEM uses to guide the robot to the target quickly.
Conversely, FUEL-complete misses the semantic cues and first explores the left arm of the cave before eventually discovering the target. FUEL-original coincidentally explores the right arm of the cave first, but does not find the target due to its thresholding of frontier clusters and viewpoint gains.

\subsection{Planner Ablation Study}
\label{sec:results_ablations}
In this section, we evaluate the importance of the combinatorial target search planner in our method. 
We compare our full method with the WMLP-based planner
from \cref{sec:method_WMLP}
with two simplified variants that use the same viewpoint sampling, evaluation, and filtering methods using semantics, but replace the planner. 
The first variant uses a greedy viewpoint choice, planning to the viewpoint with the highest information gain $I_\nu$.
The second variant uses the TSP planner from \cite{zhou2021fuel} to plan a path through the viewpoints in $\cV$, deploying the LKH heuristic to solve the optimization problem. The quantitative results are shown in \cref{tab:planning_ablation}.

\begin{table*}[t]
    \caption{Planner ablation study, comparing our proposed WMLP planner with a greedy and a TSP-based planner on success \%, time to target $t*$, and exploration time $t_f$.}
    \label{tab:planning_ablation}
    \centering
    \begin{tabular}{c|l|cc|c}
    \toprule
    \multicolumn{2}{c}{} & \multicolumn{2}{|c|}{Target search}&  Exploration\\
    \midrule
     Env & Method& Success \%  &  Time, $t^*$(s)  & Time, $t_f$(s) \\
     \midrule
     \multirow{3}{*}{Earthquake}    
         & Greedy & 100\% &  $71.1 \pm 39.1$      & $210.5 \pm 25.8$      \\
         & TSP-LKH & 100\%   &  $65.3 \pm 26.7$       & \textbf{$128.5 \pm 10.8$} \\
         & WMLP (STEM) & 100\% & \textbf{$56.9 \pm 22.8$}  & $139.9 \pm 11.9$      \\
     \midrule
      \multirow{3}{*}{Cave}    
         & Greedy & 50\%      & $64.9 \pm 18.1$  & $175.8 \pm 35.8 $     \\
         & TSP-LKH & 85\%         & $93.7 \pm 19.9$       & $134.1 \pm 23.5$ \\
         & WMLP (STEM) & 90\%       & \textbf{$64.1 \pm 20.3$}        & \textbf{$130.7 \pm 19.8$}     \\
     \bottomrule
    \end{tabular}
\end{table*}

The results show that STEM with the WMLP planner outperforms both the TSP and greedy planners on target search metrics in both environments.
In the earthquake environment, all three variants achieve a 100\% success rate, while in the cave environment, the success rate of the greedy planner is substantially lower than that of WMLP and TSP.
This indicates that the complex geometry of the cave environment requires more consistent and efficient behavior of the combinatorial planners than the simpler earthquake environment.

The average target search times $t^*$ show significant differences between the planners in both environments.
In the earthquake environment, the TSP planner is 15\% slower, and the greedy planner is 25\% slower in finding the target than the WMLP planner.
This underlines that the WMLP planner prioritizing semantically relevant viewpoints contributes substantially to STEM's target search performance.
However, it also shows that the performance gap between FUEL and STEM (\cref{sec:results_quant}) is only partially caused by the planning algorithm, and that the semantic diffusion-based viewpoint evaluation and filtering are also critical elements of the target search strategy.
In the cave environment, the greedy planner achieves equally fast search times as WMLP \textit{when successful}, while the TSP planner is 46\% slower. 
This indicates that in some cases, maximizing semantic viewpoint gains via the greedy planner helps to find the target quickly.
The greedy strategy can work well when the environment constrains the exploration options (e.g., by tunnels) and when unambiguous semantic cues are continually observed.
In contrast, the TSP planner treats all viewpoints as equally important and only optimizes for their efficient coverage.

While WMLP and TSP show very similar exploration times, the greedy planner is significantly less efficient in covering the volume.
This is due to the greedy planner's tendency to oscillate between viewpoints with high information gain, leading to long detours and inefficient exploration paths.

In summary, this ablation study shows that combinatorial planning contributes to more consistent performance for both target search and exploration, and that both the semantic viewpoint evaluation and the WMLP planner contribute substantially to the target search performance of STEM.

\section{Real-world experiments}\label{sec:results_hw}
This section presents the results of real-world experiments conducted
using the software architecture from Section \ref{sec:setup_arch} and the hardware setup from Section \ref{sec:setup_rw}. 

\begin{figure*}
    \centering
    \includegraphics[width=1\linewidth, trim=0 180 0 0, clip]{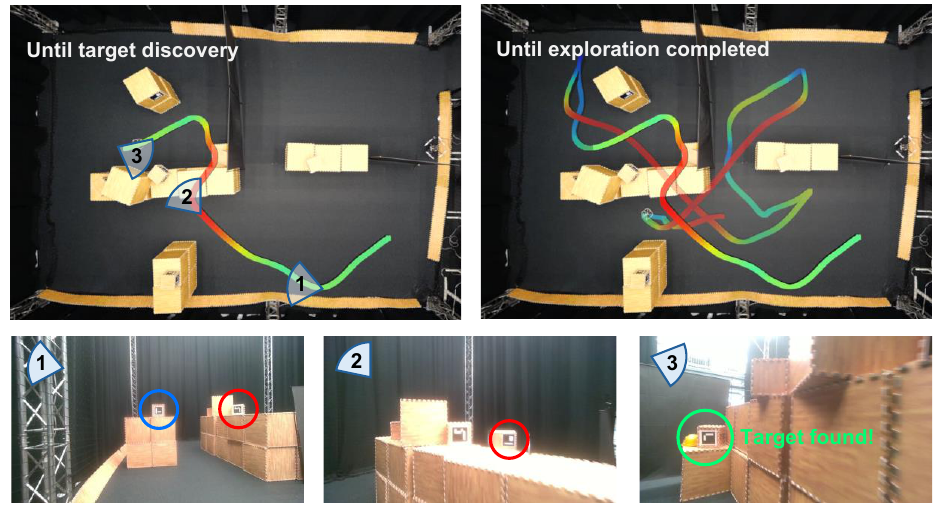}
    \caption{Hardware experiment 1 in the laboratory. The top row shows two top-down views. The color of the trajectory indicates the drone's altitude $z$. The bottom row shows images from the onboard RGB camera at three time instances, which are marked by position and orientation in the left top-down view. The red circles in the onboard images indicate the high-priority objects detected by the framework, which guide the drone towards the target. The blue circle indicates a low-priority distractor object. The last onboard image shows the moment the target is discovered.}
    \label{fig:rw_experiment1}
    \vspace{-0.5em}
\end{figure*}
\begin{figure*}
    \centering
    \includegraphics[width=1\linewidth, trim=0 180 0 0, clip]{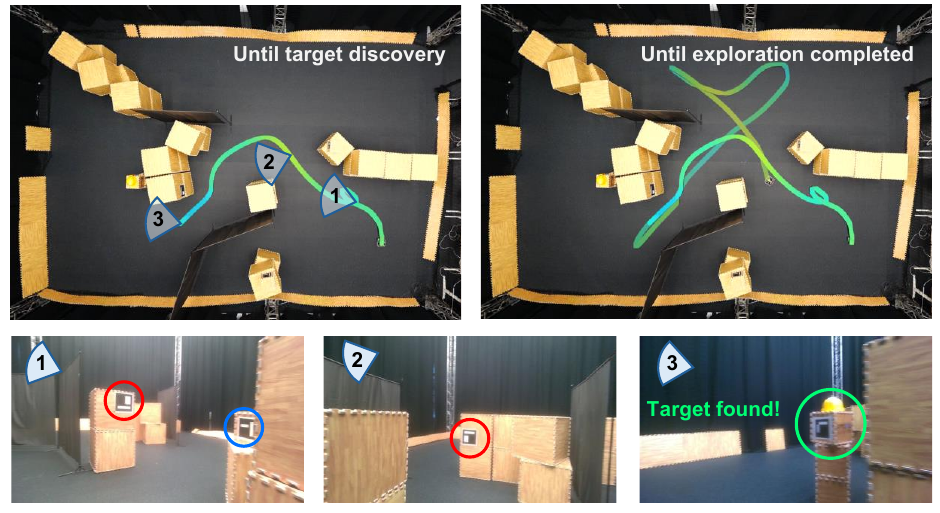}
    \caption{Hardware experiment 2 in the laboratory, with the same format as \Cref{fig:rw_experiment1}.}
    \label{fig:rw_experiment2}
\end{figure*}
\begin{figure*}
    \centering
    \includegraphics[width=1\linewidth, trim=0 180 0 0, clip]{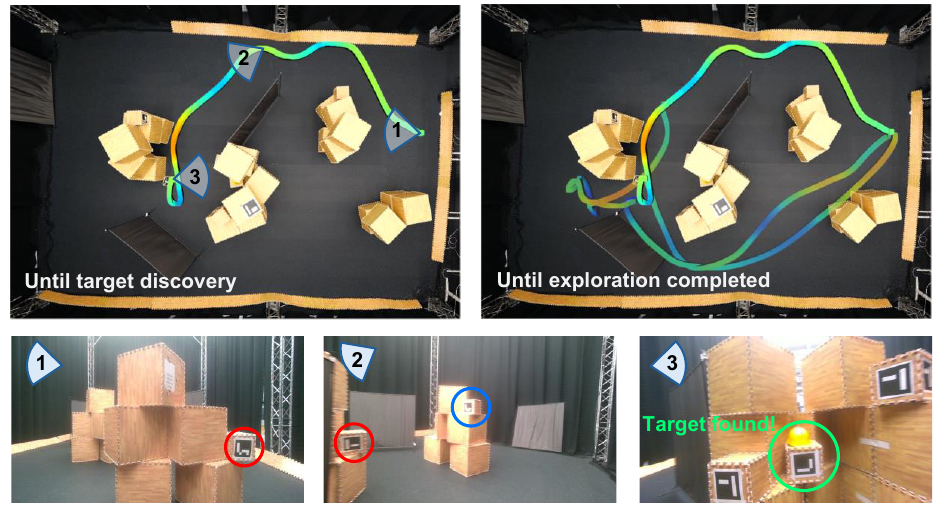}
    \caption{Hardware experiment 3 in the laboratory, with the same format as \Cref{fig:rw_experiment1}.}
    \label{fig:rw_experiment3}
\end{figure*}

The results are visualized in \Cref{fig:rw_experiment1,fig:rw_experiment2,fig:rw_experiment3}, showing our algorithm performing simultaneous target search and exploration in complex 3D environments set up in the laboratory.
\Cref{fig:rw_experiment1} shows a scenario where the drone starts from the bottom right corner in the top-down view, and explores the environment searching for the target and relevant objects, represented by ArUco markers. 
The target is hidden behind a screen and a lower wall in the upper left section of the lab. The onboard camera images in \Cref{fig:rw_experiment1} show how the drone first discovers a high-priority object on top of the wall on the right and a low-priority on the left, and therefore prioritizes the unknown area behind the wall on the right due to the diffusion method (see \cref{sec:method_frontier_diffusion}). In image 2, a high-relevance object is detected behind the wall, creating additional high-priority frontiers around the pillar in the upper left corner, which leads the WMLP planner to prioritize exploring the area behind the wall and screen further. Therefore, the drone is able to discover the target quickly and without large detours in image 3. 
Following target discovery, the framework guides the drone to further explore the environment efficiently, facilitated by the viewpoint evaluation method combining coverage and semantic information gain (see \cref{sec:method_info_gain}).
This scenario underlines how our framework is able to search for a target using semantic clues in a cluttered 3D environment, guiding the drone to fly over obstacles (the wall close to viewpoint 2) to explore important hidden areas behind it.

\Cref{fig:rw_experiment2,fig:rw_experiment3} show two additional scenarios, where the drone is able to find the target quickly using the ArUco marker objects as semantic clues, as described above. 
Both experiments show how our framework is able to guide the drone through different cluttered environments, handling occlusions and narrow passages, and to efficiently explore the environment after finding the target.

In summary, the real-world experiments demonstrate the feasibility of our method in handling noisy onboard depth images and real-time constraints, and still efficiently guide the drone towards semantically important areas. This enables the drone to find the target quickly in three different complex 3D environments, and to explore the environment efficiently after target discovery.
\section{Conclusion}
\label{sec:conclusion}
In this paper, we introduced STEM, a novel framework for semantically-guided target search and exploration with Micro Aerial Vehicles (MAVs) in complex, cluttered 3D environments. 
The core contribution of this work is the integration of efficient planning techniques commonly used in coverage exploration with semantic relationships guiding target search.
The key idea of our approach is to prioritize exploration around semantically relevant objects when planning over all available exploration frontiers.
Balancing semantic and coverage objectives in this way, efficient behavior for both target search and exploration is achieved.

Quantitative evaluation in two challenging simulation environments demonstrates that STEM achieves consistently higher success rates of more than 90 \% and up to 32\% faster target discovery times compared to coverage-maximizing baseline methods. 
Real-world experiments further validate the practical applicability of our approach, 
proving that our approach can exhibit effective target search with noisy camera inputs and realistic drone dynamics.
STEM is designed to generalize to diverse environments by balancing semantic and coverage exploration, such that the system defaults to coverage when semantic information is absent or ambiguous.
Future work will conduct extensive experiments to evaluate how our method generalizes to environments with diverse topologies and semantic features. It will also investigate how uncertainty about semantic object detection and semantic relationships can be propagated into the active perception pipeline, and how foundation models can be used to infer semantic priorities beyond embedding similarities.

\backmatter

\bmhead{Supplementary information}

A video of the drone experiments described in \cref{sec:results_hw} is submitted together with this paper.
The code used for simulation and hardware experiments is available here: \url{https://github.com/tud-amr/stem}.

\bmhead{Acknowledgements}

This work was supported by the National Police of the Netherlands. All content represents
the opinion of the authors, which is not necessarily shared or endorsed by
their respective employers and/or sponsors. 

\section*{Declarations}

\begin{itemize}
\item No conflict of interest.

\item \textbf{Author contribution:} M.L. proposed the initial idea of extending frontier-based MAV exploration with semantic priorities, developed and implemented the target search planner, and contributed to the experiments and writing of the paper.
N.S. developed and implemented the pipelines for semantic priority masking and active perception and contributed to the experiments and writing of the paper.
L.F., R.B. and J.A.-M. provided discussions and feedback on the ideas, methods, results, and the writing of the paper.
\end{itemize}

\noindent

\bibliography{main_new_clean}

\end{document}